\documentclass[conference]{IEEEtran}
\IEEEoverridecommandlockouts

\usepackage{cite}
\usepackage{amsmath,amssymb,amsfonts}
\usepackage{algorithmic}
\usepackage{graphicx}
\usepackage{textcomp}
\usepackage{xcolor}
\usepackage{makecell}
\usepackage{booktabs}
\usepackage{multirow} 
\usepackage{graphicx} 
\usepackage{subcaption}
\def\BibTeX{{\rm B\kern-.05em{\sc i\kern-.025em b}\kern-.08em
    T\kern-.1667em\lower.7ex\hbox{E}\kern-.125emX}}
\begin{document}

\title{Learning Semantic Directions for Feature Augmentation in Domain-Generalized Medical Segmentation
\thanks{\textsuperscript{*}Corresponding author.}
}


\author{
\IEEEauthorblockN{
Yingkai Wang\IEEEauthorrefmark{1}\IEEEauthorrefmark{3},
Yaoyao Zhu\IEEEauthorrefmark{2},
Xiuding Cai\IEEEauthorrefmark{1}\IEEEauthorrefmark{3},
Yuhao Xiao\IEEEauthorrefmark{1}\IEEEauthorrefmark{3},
Haotian Wu\IEEEauthorrefmark{1}\IEEEauthorrefmark{3},
Yu Yao\IEEEauthorrefmark{1}\IEEEauthorrefmark{3}\textsuperscript{*}
}

\IEEEauthorblockA{\IEEEauthorrefmark{1} Chengdu Institute of Compute Application, Chinese Academy of Sciences, Chengdu, China}
\IEEEauthorblockA{\IEEEauthorrefmark{2} China Zhenhua Research Institute Co., Ltd., Guiyang, China}
\IEEEauthorblockA{\IEEEauthorrefmark{3} School of Computer Science and Technology, University of Chinese Academy of Sciences, Beijing, China}

\IEEEauthorblockA{
Email: wangyingkai22@mails.ucas.ac.cn, zhuyaoyao19@mails.ucas.ac.cn, caixiuding20@mails.ucas.ac.cn,\\
xiaoyuhao24@mails.ucas.ac.cn, wuhaotian23@mails.ucas.ac.cn, casitmed2022@163.com
}

}

\maketitle

\maketitle

\begin{abstract}
    Medical image segmentation plays a crucial role in clinical workflows, but domain shift often leads to performance degradation when models are applied to unseen clinical domains. This challenge arises due to variations in imaging conditions, scanner types, and acquisition protocols, limiting the practical deployment of segmentation models. Unlike natural images, medical images typically exhibit consistent anatomical structures across patients, with domain-specific variations mainly caused by imaging conditions. This unique characteristic makes medical image segmentation particularly challenging.

    To address this challenge, we propose a domain generalization framework tailored for medical image segmentation. Our approach improves robustness to domain-specific variations by introducing implicit feature perturbations guided by domain statistics. Specifically, we employ a learnable semantic direction selector and a covariance-based semantic intensity sampler to modulate domain-variant features while preserving task-relevant anatomical consistency. Furthermore, we design an adaptive consistency constraint that is selectively applied only when feature adjustment leads to degraded segmentation performance. This constraint encourages the adjusted features to align with the original predictions, thereby stabilizing feature selection and improving the reliability of the segmentation.
    
    Extensive experiments on two public multi-center benchmarks show that our framework consistently outperforms existing domain generalization approaches, achieving robust and generalizable segmentation performance across diverse clinical domains.

\end{abstract}


\section{Introduction}
Medical image segmentation plays a pivotal role in a wide range of clinical workflows, enabling accurate delineation of anatomical structures and pathological regions~\cite{litjens2017survey}. With the rapid development of deep learning, both convolutional neural networks (CNNs)\cite{ronneberger2015u}~\cite{isensee2021nnu} and transformer-based architectures\cite{hatamizadeh2022unetr}~\cite{hatamizadeh2021swin} have achieved remarkable performance. However, these models often suffer from a significant performance drop when deployed on unseen domains—such as images acquired from different institutions, scanners, or imaging protocols—due to the well-known \emph{domain shift} problem~\cite{glocker2019machine}~\cite{yoon2024domain}, which poses a major challenge for real-world clinical adoption.

Medical images exhibit relatively consistent anatomical structures, whereas domain-specific variations primarily stem from imaging conditions, including differences in scanner types, acquisition parameters, and noise characteristics. This observation motivates the development of domain generalization (DG) methods, which aim to enhance model robustness to such domain-specific factors while maintaining anatomical fidelity. In contrast to domain adaptation (DA)\cite{zhang2021prototypical}~\cite{tomar2021self}~\cite{dong2021and}, which requires access to unlabeled target-domain data, DG methods~\cite{liu2020shape}~\cite{ouyang2022causality}~\cite{gao2023bayeseg}~\cite{zhu2025improving} are designed to generalize to entirely unseen domains without relying on target data, making them particularly appealing for real-world clinical applications.

Existing DG methods for medical image segmentation commonly rely on data augmentation, which can be broadly categorized into image-level and feature-level strategies. Image-level augmentations increase data diversity but are non-end-to-end and often fail to capture fine-grained semantic details. Feature-level augmentations, on the other hand, typically inject identical noise across all channels, lacking spatial or channel-wise selectivity, which can weaken the discriminative power of learned representations.

To overcome these limitations, we propose an end-to-end segmentation framework featuring \emph{implicit feature-level perturbation} guided by domain statistics. The key idea is to perturb domain-specific components of the feature representations while preserving anatomical semantics, thereby enhancing generalization without compromising task-relevant information. Our method introduces a \textbf{learnable Semantic direction selector} that adaptively identifies optimal perturbation channels, coupled with a \textbf{covariance-based Semantic Intensity Sampler} that adjusts perturbation magnitudes according to inter-domain variability.

To stabilize the learning of the direction selector, we introduce a \textbf{Selective Consistency Loss} that encourages the segmentation outputs of original and augmented features to remain consistent. Specifically, SCL imposes consistency only on augmented features that yield lower Dice Similarity Coefficient (DSC) than the original features, while ignoring those that improve segmentation. This selective mechanism stabilizes the direction selector and guides it to identify reliable channels for augmentation.

Experiments on multi-center medical segmentation benchmarks demonstrate that our method consistently outperforms existing DG approaches, achieving robust and generalizable performance across diverse unseen domains.

Our main contributions are as follows:
\begin{itemize}
    \item We propose a DG segmentation framework with structured, domain-statistics-guided feature perturbations.  
    \item We introduce a learnable semantic direction selector combined with a covariance-based semantic intensity sampler for adaptive feature perturbation.  
    \item We design a Selective Consistency Loss that selectively regularizes degraded augmented features, improving robustness and generalization.  
\end{itemize}

\section{Related Work}
\label{sec:related_work}

\subsection{Domain Generalization in Medical Image Segmentation}

In medical image segmentation, DG aims to improve model robustness on unseen target domains where target data are unavailable during training. This is particularly critical in clinical practice due to data heterogeneity caused by variations across hospitals, imaging devices, scanning protocols, and patient populations. Among existing DG approaches, augmentation-based methods are especially appealing for their simplicity and effectiveness, which motivates our focus on augmentation-centric DG frameworks. Such methods can be broadly categorized into image-level~\cite{billot2023synthseg}\cite{shen2022randstainna}\cite{scalbert2022test} and feature-level augmentations~\cite{meng2020mutual}\cite{bi2023mi}. By leveraging these augmentations, DG methods can better simulate domain shifts and learn representations that remain robust to variations in both visual appearance and semantic features.

\subsubsection{Image-Level Augmentation}

Image-level augmentation is a classical DG strategy to simulate domain shifts, alleviate data scarcity, and enhance cross-domain adaptability. 
Typical techniques include random flipping, rotation, scaling, and cropping. 
Ouyang et al.~\cite{ouyang2022causality} reduced spurious correlations via diverse appearance augmentations and causal interventions, improving robustness across devices. 
Zhou et al.~\cite{zhou2022generalizable} proposed a B\'ezier-curve-based image generation method to produce both source-similar and source-dissimilar samples, 
combined with a dual-normalization network to boost generalizability. 
However, most image-level approaches operate in the global image space, are not fully end-to-end, and often rely on task-specific augmentation policies, 
limiting the diversity of generated data.

\subsubsection{Feature-Level Augmentation}

Feature-level augmentation enables more flexible and semantically rich transformations in the latent space compared to image-level operations. 
Chen et al.~\cite{chen2023treasure} expanded the feature style space through statistical randomization and introduced a fusion strategy to learn domain-invariant representations. 
Zhang et al.~\cite{zhang2024domain} preserved inter-feature correlations to achieve style augmentations beyond simple interpolation, producing more diverse and semantically consistent transformations. 
While effective for improving generalization, these methods may disrupt critical structures such as lesion boundaries in the absence of semantic guidance, potentially compromising segmentation accuracy.

\section{Methods}

\begin{figure*}[ht]
    \centerline{\includegraphics[width=0.8\textwidth]{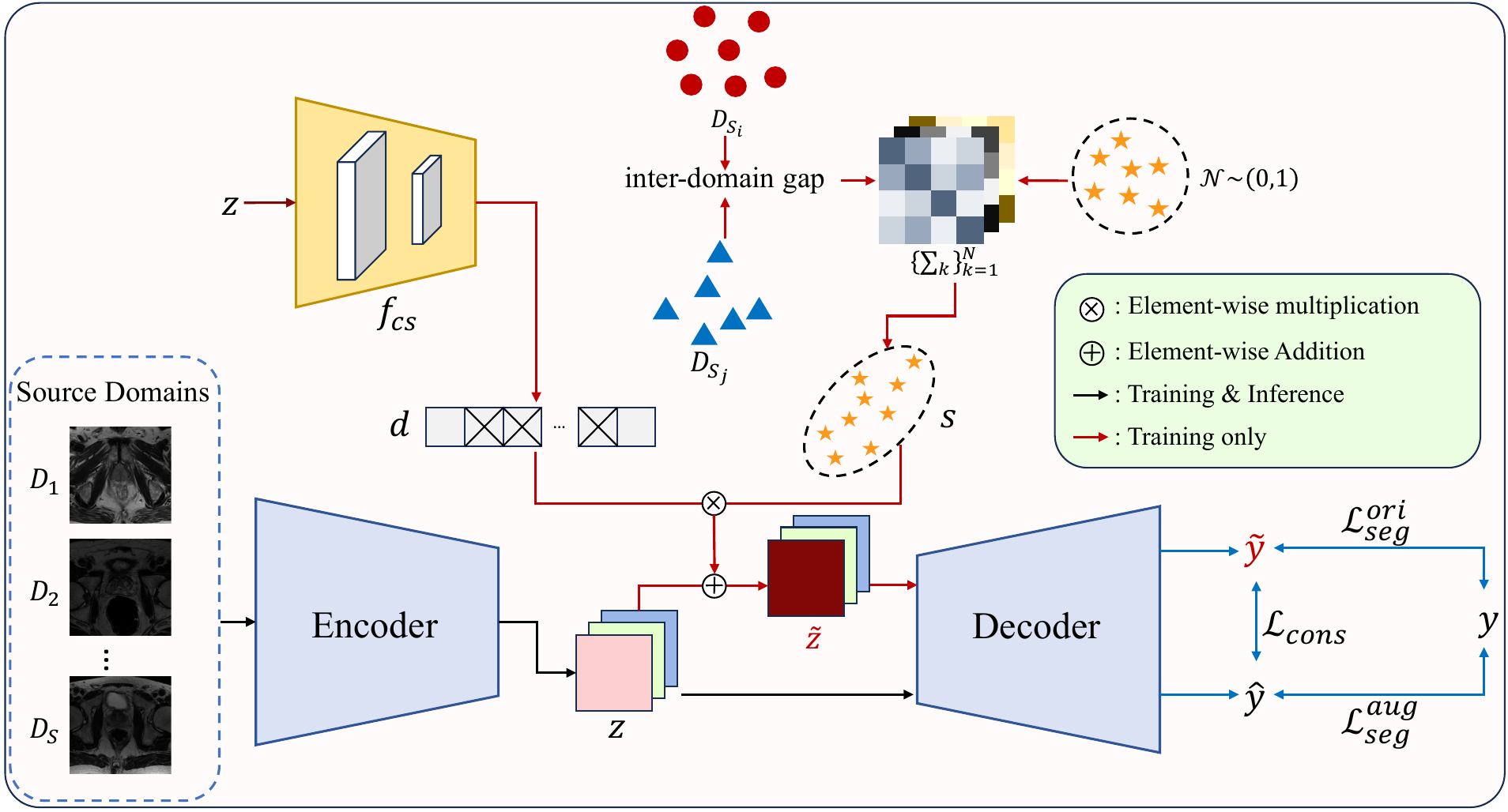}}
    \caption{Overview of the proposed Semantic Direction Feature Augmentation (SDFA) framework.
    The model is trained on multiple source domain datasets to learn generalizable representations for segmentation on unseen target domains.}
    \label{main}
    \end{figure*}

\subsection{Preliminaries}
\subsubsection{Problem Setting}
Assume that we have $K$ source domains collected from different medical centers, denoted as:
$
D^S = \{D_1^S, D_2^S, \dots, D_K^S\}
$
In the $k$-th source domain $D_k^S$, the training samples and their corresponding segmentation masks can be represented as:
$
D_k^S = \left\{ (\mathbf{x}_{k,n}^S, \mathbf{y}_{k,n}^S) \right\}_{n=1}^{N_k}
$
where $\mathbf{x}_{k,n}^S$ denotes the $n$-th image in the $k$-th source domain, $\mathbf{y}_{k,n}^S$ is the corresponding segmentation mask, and $N_k$ is the number of samples in domain $D_k^S$.

In the domain-generalized medical image segmentation task, our goal is to train a robust segmentation model $F_\theta$ using the source domain data $D^S$, such that it can generalize well to the unseen target domain(s), denoted as:
$
D^T = \left\{ \mathbf{x}_n^T \right\}_{n=1}^{N_T}
$
where $\mathbf{x}_n^T$ is the $n$-th sample in the target domain, and $N_T$ is the number of target samples.

\subsubsection{Semantic Data Augmentation}
Deep learning models are known to linearize semantic representations~\cite{upchurch2017deep}, enabling the generation of semantically diverse features through linear perturbations while preserving the original labels~\cite{wang2021regularizing}. Zhu \textit{et al.} ~\cite{zhu2024bsda} redefined semantic data augmentation by decomposing the process into two core components: \textit{semantic direction} and \textit{semantic intensity}. Specifically, the semantic direction determines which feature channels are to be altered to enable semantic transformation, while the semantic intensity modulates the magnitude of this transformation, as excessive perturbation may alter the semantic label.

This study introduces a semantic data augmentation module specifically designed for DG in medical image segmentation. The formulation of the proposed module is as follows:
\begin{equation}
\tilde{\mathbf{z}}_i = \mathbf{z}_i + \mathbf{d} \odot \mathbf{s},
\end{equation}
where $\mathbf{z}_i = \boldsymbol{\Phi}(\mathbf{x}_i)$ represents the deep feature extracted by the encoder for input $\mathbf{x}_i$, $\mathbf{d}$ is the learned semantic direction vector, $\mathbf{s}$ is the sampled semantic intensity vector, and $\odot$ denotes element-wise multiplication.

\subsection{Semantic Data Augmentation Framework}
\subsubsection{Semantic Direction Selector (SDS)}

Let the encoded feature map be denoted as $\mathbf{z} \in \mathbb{R}^{C \times H \times W}$, where $C$ is the number of channels and $H \times W$ represents the spatial resolution. To enhance the model's adaptability to unseen domains, we aim to selectively perturb this feature map along semantically meaningful directions.

While the covariance matrix of the features can capture inter-channel relationships~\cite{zhang2024domain}, directly using statistical measures alone to guide augmentation may introduce noise and generate semantically inconsistent features. 
To address this issue, we design a learnable semantic direction selector function $f_{\mathrm{cs}}$, which consists of two convolutional layers, followed by a global average pooling and a fully connected layer. The function outputs a channel-wise vector:
\begin{equation}
    \mathbf{d} = f_{\mathrm{cs}}(\mathbf{z}), \mathbf{d} \in \{0,1\}^C,
    \end{equation} 
Each element $\mathbf{d_c}$ indicates whether channel $\mathbf{c}$ is selected for semantic augmentation.
Specifically, $\mathbf{d_c} = 1$ denotes that channel $\mathbf{c}$ is deemed suitable for augmentation, whereas $\mathbf{d_c} = 0$ indicates that the channel should remain unchanged.

This learnable mechanism allows the model to adaptively select augmentation directions along task-relevant semantic axes, thereby enhancing cross-domain generalization without compromising semantic integrity.

\subsubsection{Semantic Intensity Sampler (SIS)}

Given the semantic direction $\mathbf{d}$, we introduce a semantic intensity variable $\mathbf{s}$ sampled from a Gaussian distribution $\mathbf{s} \sim \mathcal{N}(0, 1)$ to control the strength of augmentation along the selected semantic dimensions. However, a fixed distribution may lead to suboptimal augmentation—either too weak or too strong—thus compromising generalization performance. 

To control the strength of semantic augmentation, we introduce a semantic intensity variable. Instead of using a fixed Gaussian prior, we design the SIS guided by inter-domain covariance statistics. Specifically, we compute a set of covariance matrices $\{\Sigma_k\}_{k=1}^N$, where each $\Sigma_k \in \mathbb{R}^{C \times C}$ captures the inter-domain covariance between a pair of domains $(\mathcal{D}_{s_i}, \mathcal{D}_{s_j})$:
\begin{equation}
\Sigma_k = \mathrm{Cov}(\mathbf{z}_{s_i}, \mathbf{z}_{s_j}) = \mathbb{E}[(\mathbf{z}_{s_i} - \bar{\mathbf{z}}_{s_i})(\mathbf{z}_{s_j} - \bar{\mathbf{z}}_{s_j})^\top].
\end{equation}
For a feature $\mathbf{z}$ to be augmented, we randomly select another domain to form a domain pair and sample semantic noise from the corresponding covariance:
\begin{equation}
\boldsymbol{\xi} \sim \mathcal{N}(0, \Sigma_k).
\end{equation}

However, the inter-domain covariance can only reflect the response differences across channels between two domains; it cannot directly and reliably determine how much augmentation should be applied to a specific feature instance from a particular domain. To achieve instance-adaptive augmentation, we introduce a pair of learnable parameters, $\boldsymbol{\mu} \in \mathbb{R}^C$ and $\boldsymbol{\sigma} \in \mathbb{R}^C$, which control the shift and scaling of the semantic intensity. The final adaptive semantic intensity is given by:
\begin{equation}
\mathbf{s} = \boldsymbol{\mu} + \boldsymbol{\sigma} \odot \boldsymbol{\xi},
\end{equation}

This instance-adaptive scaling allows the augmentation strength to reflect domain-specific feature variability, improving the generalization to unseen domains.

\subsection{Optimization Objective Design}

We design two loss functions to optimize our model: the supervised segmentation loss and the selective consistency loss.

\subsubsection*{Supervised Segmentation Loss}

Given a labeled sample $\mathbf{x}_i$ with ground-truth mask $\mathbf{y}_i$, 
the supervised segmentation loss is formulated as a combination of the losses 
computed from the prediction of the original input and the prediction of the augmented features:
\begin{equation}
\mathcal{L}_{\text{sup}}
= \mathcal{L}_{\text{seg}}^{\text{ori}} 
+ \lambda \cdot \mathcal{L}_{\text{seg}}^{\text{aug}},
\label{eq:supervised_seg}
\end{equation}
where
\begin{align}
\mathcal{L}_{\text{seg}}^{\text{ori}} 
&= \mathcal{L}_{\text{Dice}}\big(\hat{\mathbf{y}}_i, \mathbf{y}_i\big)
+ \mathcal{L}_{\text{CE}}\big(\hat{\mathbf{y}}_i, \mathbf{y}_i\big), \\
\mathcal{L}_{\text{seg}}^{\text{aug}} 
&= \mathcal{L}_{\text{Dice}}\big(\tilde{\mathbf{y}}_i, \mathbf{y}_i\big)
+ \mathcal{L}_{\text{CE}}\big(\tilde{\mathbf{y}}_i, \mathbf{y}_i\big).
\end{align}

In this formulation, $\hat{\mathbf{y}}_i$ denotes the segmentation result 
obtained from the original feature $\mathbf{z}_i$, 
while $\tilde{\mathbf{y}}_i$ represents the result derived from the augmented feature $\tilde{\mathbf{z}}_i$. 
The hyperparameter $\lambda$ balances the contribution of the augmented loss.

\begin{table*}[!t]
    \centering
    \caption{Detailed information of Fundus and Prostate datasets}
    \label{tab:datasets}
    \begin{tabular}{llcccccc}
    \toprule
    Datasets & Configure & Domain-1 & Domain-2 & Domain-3 & Domain-4 & Domain-5 & Domain-6 \\
    \midrule
    \multirow{3}{*}{Fundus} 
    & Sub-dataset  & Drishti-GS        & RIM-ONE-r3         & REFUGE (train)  & REFUGE (val.) & /    & / \\
    & Vendor       & Aravind eye hosp. & Nidek AFC-210      & Zeiss Visucam 500 & Canon CR-2 & /    & / \\
    & Data no. & 101             & 159             & 400          & 400      & /    & / \\
    \midrule
    \multirow{3}{*}{Prostate}
    & Inst.        & RUNMC             & BMC                & HCRUDB          & UCL          & BIDMC & HK \\
    & Vendor       & Siemens           & Philips            & Siemens         & Siemens      & GE    & Siemens \\
    & Case no.     & 30                & 30                 & 19              & 13           & 12    & 12 \\
    \bottomrule
    \end{tabular}
    \label{dataset details}
    \end{table*}

\subsubsection*{Selective Consistency Loss(SCL)}
To stabilize the learning of the SDS, 
we introduce SCL, 
which encourages the enhanced features to maintain or improve segmentation performance. 
For each sample, we compare the segmentation losses of the original and enhanced features. 
If the enhanced features yield a higher segmentation loss 
(i.e., $\mathcal{L}_{\text{seg}}^{\text{aug}} > \mathcal{L}_{\text{seg}}^{\text{ori}}$), 
the enhancement is considered harmful, and a penalty is applied; 
otherwise, no penalty is imposed. 
Formally, SCL is defined as:

\begin{equation}
\mathcal{L}_{\text{cons}} = \frac{1}{N} \sum_{i=1}^{N} \mathbb{I}(\mathcal{L}_{\text{seg}}^{\text{aug}, i} > \mathcal{L}_{\text{seg}}^{\text{ori}, i}) \cdot \left| \mathcal{L}_{\text{seg}}^{\text{aug}, i} - \mathcal{L}_{\text{seg}}^{\text{ori}, i} \right|,
\end{equation}
where $\mathbb{I}(\cdot)$ denotes the indicator function, and $N$ is the batch size.
This design enforces that feature augmentations should either maintain or improve segmentation performance,
while only harmful augmentations are penalized, thereby stabilizing the learning of the SDS.

\subsubsection*{Total Loss}
The overall training loss consists of the supervised segmentation loss and the SCL:
\begin{equation}
\mathcal{L}_{\text{total}} = \mathcal{L}_{\text{seg}} +  \mathcal{L}_{\text{cons}},
\end{equation}

\section{EXPERIMENTS}
\subsection{Datasets and Evaluation Metrics}

We evaluate our method on two publicly available multi-center medical image segmentation benchmark datasets: the DG Prostate benchmark~\cite{liu2020shape} and the DG Fundus benchmark~\cite{wang2020dofe}. These datasets are particularly suitable for assessing the generalization ability of segmentation models across different domains. Detailed information about the datasets is provided in Table~\ref{dataset details}.

For the Prostate segmentation task, following previous works~\cite{chen2023treasure,bi2024learning}, we convert the original 3D volumes into 2D slices for both training and evaluation. During training, the input size for the Fundus dataset is $512 \times 512 \times 3$, while for the Prostate dataset it is $384 \times 384 \times 1$.

To enhance model robustness, we apply standard data augmentation strategies to all methods, including brightness adjustment, contrast variation, Gaussian noise injection, and elastic deformation.

\begin{itemize}
    \item \textbf{DG Prostate Benchmark}~\cite{liu2020shape}: This dataset consists of 116 T2-weighted MRI cases collected from six hospitals, denoted as Domain-1 to Domain-6. 
    \item \textbf{DG Fundus Benchmark}~\cite{wang2020dofe}: This dataset comprises four optic disc/cup segmentation sub-datasets, namely DrishtiGS~\cite{sivaswamy2015comprehensive}, RIM-ONE-r3~\cite{fumero2011rim}, REFUGE (train)~\cite{orlando2020refuge}, and REFUGE (val)~\cite{orlando2020refuge}, denoted as Domain-1, Domain-2, Domain-3, Domain-4.
\end{itemize}

In all segmentation tasks, we adopt the Leave-One-Domain-Out evaluation protocol to measure cross-domain generalization performance. In each experiment, one domain is held out for testing, while the remaining domains are used for training and validation. Specifically, we randomly sample 20\% of data from each training domain to form the validation set for model selection and hyperparameter tuning. This setup ensures scientific rigor and prevents overfitting, while maintaining a fair evaluation since the test domain is never seen during training or validation.

We adopt two widely used evaluation metrics: Dice Similarity Coefficient (DSC) and Average Surface Distance (ASD). 
DSC quantifies the region overlap between the predicted segmentation and the ground-truth mask, 
where higher values indicate better segmentation accuracy. 
In contrast, ASD measures the average distance between the predicted and ground-truth boundaries, 
with lower values indicating more precise boundary alignment.

These metrics, combined with strict validation protocols, provide a comprehensive and reliable assessment of model performance and guide further optimization.

\begin{table*}[ht]
    \centering
    \caption{Performance comparison (DSC ↑ and ASD ↓) of the proposed method and existing methods on domain generalized Prostate Dataset segmentation task. Average indicates the mean DSC and ASD across all four domains. The best results are highlighted in bold, while the second-best results are underlined.}
        \small
        \resizebox{\textwidth}{!}{
        \begin{tabular}{lcccccccccccccc}
        \toprule
        \multirow{2}{*}{Method} & \multicolumn{2}{c}{Domain-1} & \multicolumn{2}{c}{Domain-2} & \multicolumn{2}{c}{Domain-3} & \multicolumn{2}{c}{Domain-4} & \multicolumn{2}{c}{Domain-5} & \multicolumn{2}{c}{Domain-6} & \multicolumn{2}{c}{Average} \\
        & DSC$\uparrow$ & ASD$\downarrow$ & DSC$\uparrow$ & ASD$\downarrow$ & DSC$\uparrow$ & ASD$\downarrow$ & DSC$\uparrow$ & ASD$\downarrow$ & DSC$\uparrow$ & ASD$\downarrow$ & DSC$\uparrow$ & ASD$\downarrow$ & DSC$\uparrow$ & ASD$\downarrow$ \\
        \midrule
        ERM & 86.27 & 1.97 & 83.27 & 2.24 & 87.39 & 1.81 & 85.37 & 1.58 & 78.24 & 3.13 & 84.25 & 1.71 & 84.13 & 2.07 \\
        MixStyle(ICLR2021) & 85.85 & 2.01 & 82.73 & 2.32 & 87.39 & 1.77 & 83.89 & 1.69 & 81.06 & 2.61 & 84.10 & 1.72 & 84.17 & 2.02 \\
        EFDMixStyle(CVPR2022) & 86.00 & 1.99 & 82.38 & 2.40 & 87.59 & 1.70 & 84.12 & 1.64 & 80.07 & 2.99 & 84.96 & 1.67 & 84.19 & 2.06 \\
        Causality(TMI2022) & 85.44 & 2.00 & 84.19 & 2.12 & 88.25 & 1.64 & 84.80 & 1.56 & 83.88 & 2.17 & 84.19 & 1.73 & 85.13 & 1.87 \\
        Bezier(CVPR2022)& 86.56 & 1.85 & \textbf{85.02} & \textbf{1.88} & 87.35 & 1.58 & 83.31 & 1.69 & \textbf{84.20} & \textbf{2.19} & 86.75 & 1.51 & 85.53 & 1.78 \\
        TriD(MICCAI2023) & 86.11 & 2.01 & 83.27 & 2.29 & 87.00 & 1.86 & 83.82 & 1.76 & 81.75 & 2.73 & 85.83 & 1.54 & 84.63 & 2.03 \\
        CSU(WACV2024) & 86.05 & 1.96 & 83.95 & 2.29 & 87.50 & 1.80 & 84.12 & 1.63 & 82.30 & 2.64 & 85.70 & 1.69 & 84.94 & 2.00 \\
        SDFA(ours) & \textbf{87.86} & \textbf{1.63} & 85.25 & 1.96 & \textbf{88.92} & \textbf{1.49} & \textbf{85.96} & \textbf{1.41} & 82.24 & 2.33 & \textbf{87.82} & \textbf{1.31} & \textbf{86.34} & \textbf{1.69} \\
        \bottomrule
        \end{tabular}
        }
        \label{table Prostate comparison}
\end{table*}
\subsubsection{Quantitative Results on the Fundus Dataset}
\begin{table*}[ht]
    \centering
    \caption{Performance comparison (DSC ↑ and ASD ↓) of the proposed method and existing methods on domain generalized Optic Cup segmentation task.}
    \small
    \begin{tabular}{lcccccccccc}
    \toprule
    \multirow{2}{*}{Method} & \multicolumn{2}{c}{Domain-1} & \multicolumn{2}{c}{Domain-2} & \multicolumn{2}{c}{Domain-3} & \multicolumn{2}{c}{Domain-4} & \multicolumn{2}{c}{Average} \\
    & DSC$\uparrow$ & ASD$\downarrow$ & DSC$\uparrow$ & ASD$\downarrow$ & DSC$\uparrow$ & ASD$\downarrow$ & DSC$\uparrow$ & ASD$\downarrow$ & DSC$\uparrow$ & ASD$\downarrow$ \\
    \midrule
    ERM  & 78.31 & 14.61 & 69.40 & 13.25 & 82.16 & 7.72 & 74.07 & 15.09 & 75.99 & 12.67 \\
    MixStyle(ICLR2021)  & 79.90 & 13.57 & 68.77 & \textbf{12.43} & \underline{83.48} & \underline{7.29} & 79.55 & \underline{10.22} & 77.92 & 10.88 \\
    EFDMixStyle(CVPR2022)  & 74.25 & 16.64 & \textbf{73.18} & 12.84 & 82.69 & 7.91 & 73.24 & 27.53 & 75.84 & 16.23 \\
    Causality(TMI2022) & 75.13 & 15.30 & 56.71 & 22.13 & 77.93 & 10.72 & 75.19 & 11.21 & 71.24 & 14.84 \\
    Bezier(CVPR2022)  & 78.71 & 13.65 & 61.63 & 20.95 & 82.71 & 7.70 & 64.62 & 31.84 & 71.92 & 18.54 \\
    TriD(MICCAI2023)  & 78.47 & 14.70 & 68.50 & 17.29 & 80.81 & 8.39 & \textbf{84.21} & 11.14 & 77.99 & 12.88 \\
    CSU(WACV2024)  & \textbf{80.48} & \textbf{13.30} & 72.45 & 14.49 & 81.34 & 8.10 & 76.37 & 13.04 & 77.66 & 12.23 \\
    SDFA(ours)  & \underline{80.40} & \underline{13.41} & \underline{72.54} & \underline{12.57} & \textbf{83.58} & \textbf{7.29} & \underline{81.94} & \textbf{7.72} & \textbf{79.62} & \textbf{10.25} \\
    \bottomrule
    
    \end{tabular}
    \label{table oc comparison}
    \end{table*}

\begin{table*}[htbp]
    \centering
    \caption{Performance comparison (DSC ↑ and ASD ↓) of the proposed method and existing methods on domain generalized Optic Disc segmentation task.}
    \small
    \begin{tabular}{lcccccccccc}
    \toprule
    \multirow{2}{*}{Method} & \multicolumn{2}{c}{Domain-1} & \multicolumn{2}{c}{Domain-2} & \multicolumn{2}{c}{Domain-3} & \multicolumn{2}{c}{Domain-4} & \multicolumn{2}{c}{Average} \\
    & DSC$\uparrow$ & ASD$\downarrow$ & DSC$\uparrow$ & ASD$\downarrow$ & DSC$\uparrow$ & ASD$\downarrow$ & DSC$\uparrow$ & ASD$\downarrow$ & DSC$\uparrow$ & ASD$\downarrow$ \\
    \midrule
    ERM & 93.08 & 7.74 & 83.68 & 16.85 & 92.15 & 7.23 & 90.77 & 10.31 & 89.92 & 10.53 \\
    MixStyle(ICLR2021)  & 94.15 & \underline{6.57} & \underline{85.16} & \underline{15.41} & 90.82 & 8.41 & 92.75 & 9.51 & 90.72 & 9.97 \\
    EFDMixStyle(CVPR2022)  & \underline{94.27} & 6.58 & 84.59 & 15.67 & \textbf{93.19} & \textbf{6.31} & \underline{93.35} & \underline{5.85} & 91.35 & 8.60 \\
    Causality(TMI2022)  & 92.68 & 8.31 & 75.98 & 22.32 & 91.54 & 7.75 & 91.80 & 6.31 & 88.00 & 11.17 \\
    Bezier(CVPR2022)  & 91.24 & 9.99 & 80.53 & 18.93 & 92.16 & 7.15 & 91.56 & 7.40 & 88.87 & 10.87 \\
    TriD(MICCAI2023)  & 94.10 & \textbf{6.51} & 83.88 & 19.73 & 90.16 & 8.98 & 93.25 & 8.12 & 90.35 & 10.84 \\
    CSU(WACV2024)  & 93.75 & 6.68 & 82.95 & 17.85 & 91.31 & 7.80 & 93.17 & 6.04 & 90.30 & 9.59 \\
    SDFA(ours)  & \textbf{94.34} & 6.77 & \textbf{87.23} & \textbf{12.41} & \underline{92.16} & \underline{7.14} & \textbf{94.26} & \textbf{5.43} & \textbf{92.00} & \textbf{7.94} \\
    
    \bottomrule
    \end{tabular}
    \label{table od comparison}
    \end{table*}

\subsection{Network Architecture and Training Configuration}
We adopt U-Net~\cite{ronneberger2015u} as the backbone segmentation network, which is widely used in medical image segmentation tasks. All experiments are conducted on a single NVIDIA RTX 4090 GPU. Each model is trained for 15,000 iterations using the AdamW optimizer with an initial learning rate of $3 \times 10^{-4}$ and a weight decay of 0.01 for effective parameter optimization.

\subsection{Comparison with other Methods}
\subsubsection{Quantitative Results on the Prostate Dataset}
We compare our method with the empirical risk minimization (ERM) baseline and several state-of-the-art DG approaches, including image-level data augmentation-based methods~\cite{zhou2021domain, zhang2022exact, chen2023treasure, zhang2024domain}, as well as feature-level data augmentation-based methods~\cite{ouyang2022causality, zhou2022generalizable}. 

As shown in Table~\ref{table Prostate comparison}, our method achieves the best average performance among all compared DG approaches.
Specifically, our SDFA method achieves an average DSC of 86.34, surpassing all other methods, including the widely adopted Bezier method. Although Bezier performs well on Magnetic Resonance Imaging (MRI) images, it still falls short of our approach in this task, with average DSC of 85.53.
In terms of ASD, SDFA also demonstrates a significant improvement, achieving a lower value of 1.78, compared to Causality (1.87) and CSU (2.00). These results clearly highlight the superiority of our SDFA approach in terms of both segmentation accuracy and robustness across multiple domains.

\begin{figure*}[ht]
    \centerline{\includegraphics[width=0.8\textwidth]{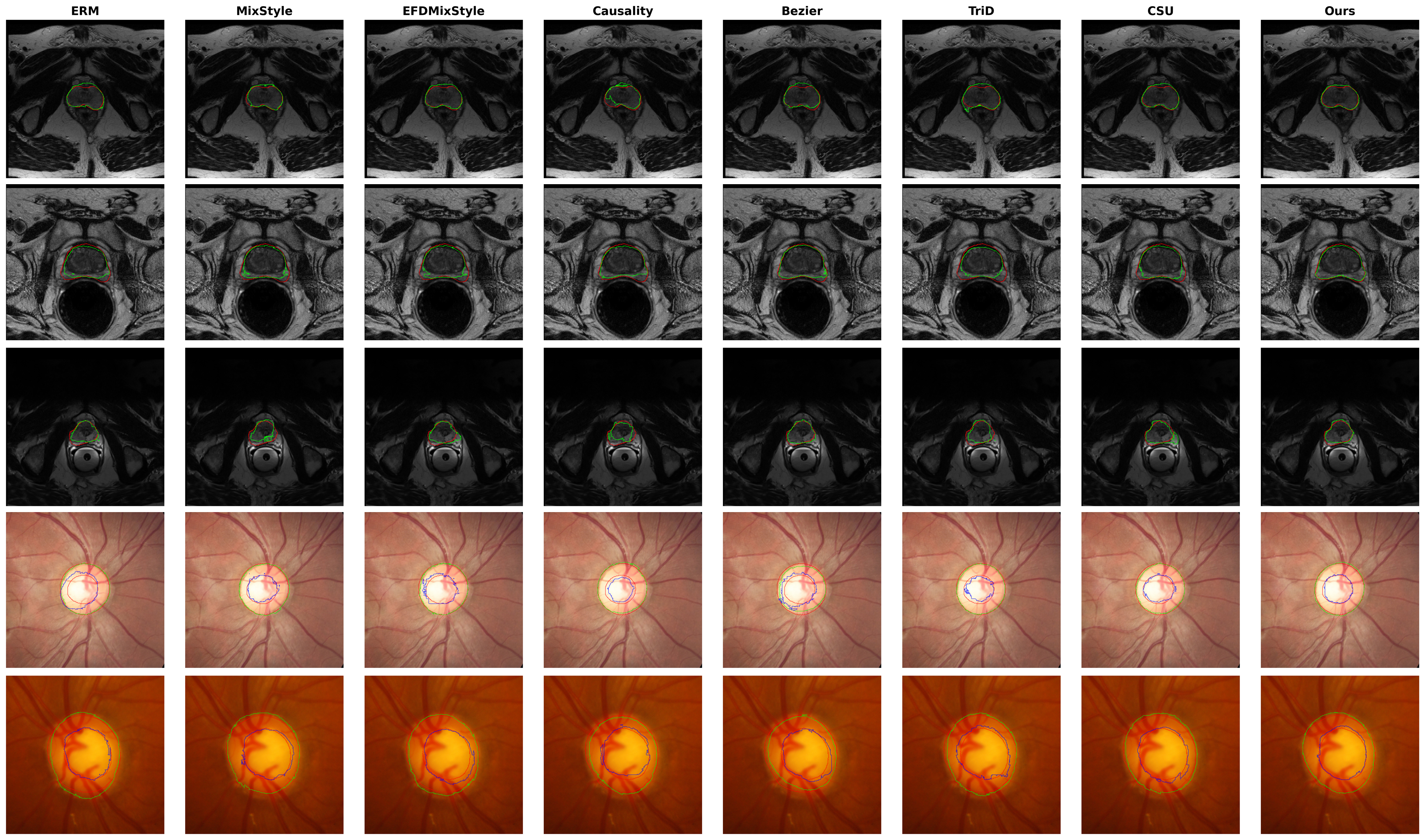}}
    \caption{Qualitative results on the Prostate and Fundus Datasets}
    \label{viz}
\end{figure*}

To further validate the effectiveness of our approach, we conducted cross-domain evaluations on optic disc and optic cup segmentation in Fundus dataset, using the same set of DG methods as in the prostate segmentation experiments. The experiments cover four different domains, and the results are summarized in Table~\ref{table oc comparison} and Table~\ref{table od comparison}.
For the optic disc task, SDFA achieves the best performance with an average DSC of 92.00 and an ASD of 7.94. For the optic cup task, SDFA also leads or closely matches the best results across most metrics. Notably, image-level augmentation methods such as Bezier perform poorly on both tasks, suggesting that such strategies may be effective only for specific imaging modalities and lack generalization to diverse Fundus dataset. These results further confirm the superior performance and robustness of our method on the Fundus dataset.

\subsubsection{Qualitative Results}
The visualization results on different datasets are presented in Figure~\ref{viz}. For the Fundus dataset, the red contours represent the ground truth, while the blue and green contours indicate the predicted boundaries of the optic cup and optic disc, respectively. For the Prostate dataset, the red contours denote the ground truth, and the green contours correspond to the predicted boundaries of the prostate. It is evident that our method delineates the boundaries in images from unseen domains more accurately compared to other approaches. Notably, our method maintains excellent boundary delineation even when the target regions are relatively large or small, whereas other methods often struggle under these challenging conditions.

\subsection{model analysis}
\subsubsection{Impact of Hyper-Parameter $\lambda$}
We introduce a hyperparameter $\lambda$ in the loss function to control the contribution of augmented features to model optimization. Table~\ref{lambda} shows a sensitivity analysis on the Prostate dataset, where $\lambda$ varies from 0.2 to 1.0 and DSC is used as the evaluation metric.

The results show that the DSC increases as $\lambda$ grows, indicating that the model benefits from the stronger supervision signal provided by the augmented features. The highest DSC of 86.34 is achieved when $\lambda = 1.0$. In general, overly aggressive data augmentation may introduce noise, and assigning high weights to such signals during optimization can lead to performance degradation. However, our results show no such decline as $\lambda$ increases, suggesting that our feature-level augmentation introduces negligible noise while still providing strong supervision. This further confirms that SDS reliably guides the augmentation process toward safe and effective directions in the feature space, ensuring that the enhanced features contribute positively to model optimization without introducing disruptive variations.
\begin{table}[htbp]
\centering
\caption{DSC for different lambda settings in domains A-F.}
\begin{tabular}{llcccccccc}
\toprule
$\lambda$ & A & B & C & D & E & F & Avg \\
\midrule

0.2 & 87.64 & 84.81 & 87.96 & 85.51 & 80.22 & 86.90 & 85.51 \\
0.4 & \textbf{87.93} & \textbf{85.36} & 87.57 & 85.76 & 80.27 & 86.06 & 85.49 \\
0.6 & 86.76 & 84.65 & 88.27 & \textbf{86.40} & 81.30 & 87.76 & 85.86 \\
0.8 & 87.49 & 84.80 & 88.68 & 85.47 & 81.88 & 87.36 & 85.94 \\
1 & 87.86 & 85.25 & \textbf{88.92} & 85.96 & \textbf{82.24} & \textbf{87.82} & \textbf{86.34} \\

\bottomrule
\end{tabular}
\label{lambda}
\end{table}

\begin{figure}[!b]
    \centering
    \begin{subfigure}[b]{0.8\columnwidth}
        \includegraphics[width=\linewidth]{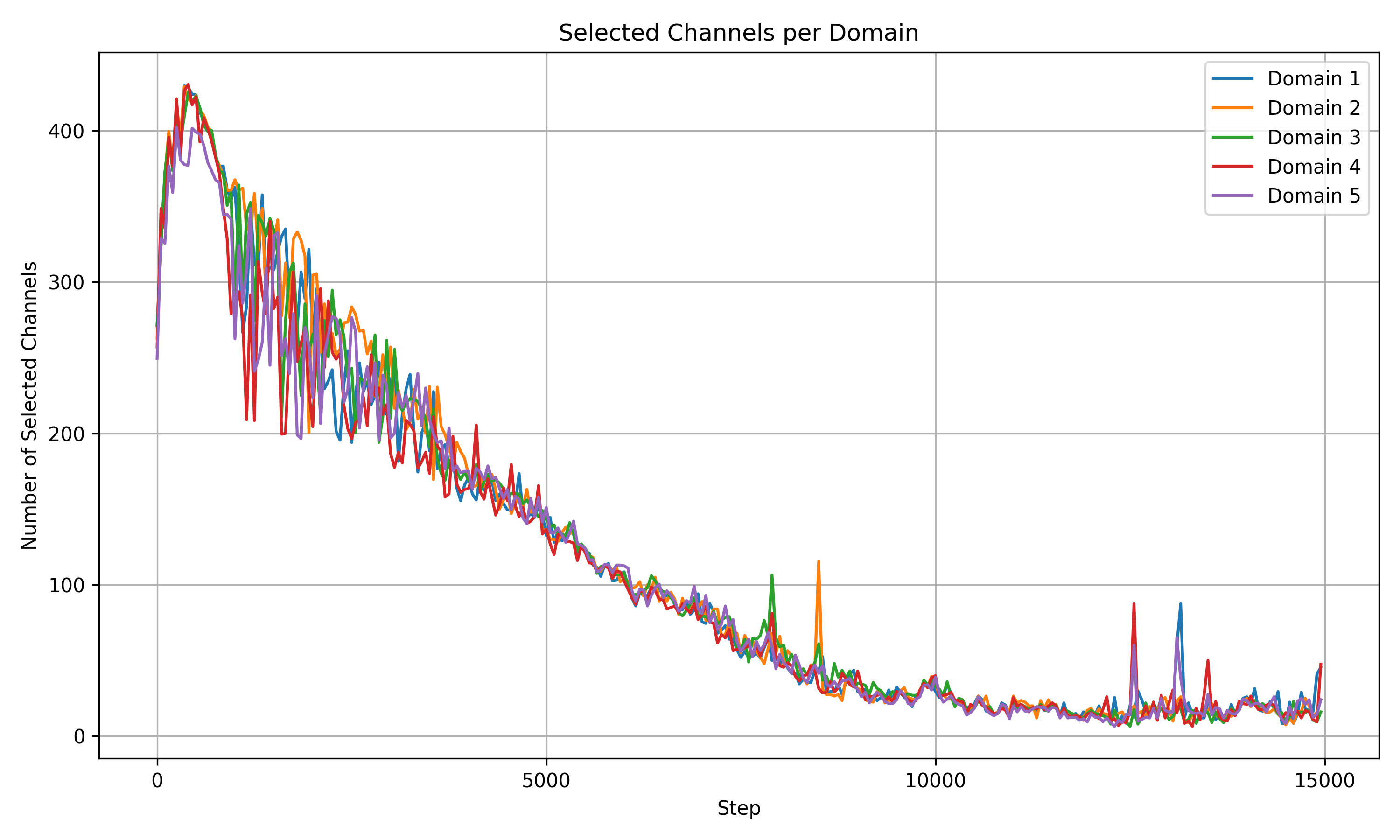}
        \caption{Without consistency loss}
    \end{subfigure}
    \vspace{1mm}
    \begin{subfigure}[b]{0.8\columnwidth}
        \includegraphics[width=\linewidth]{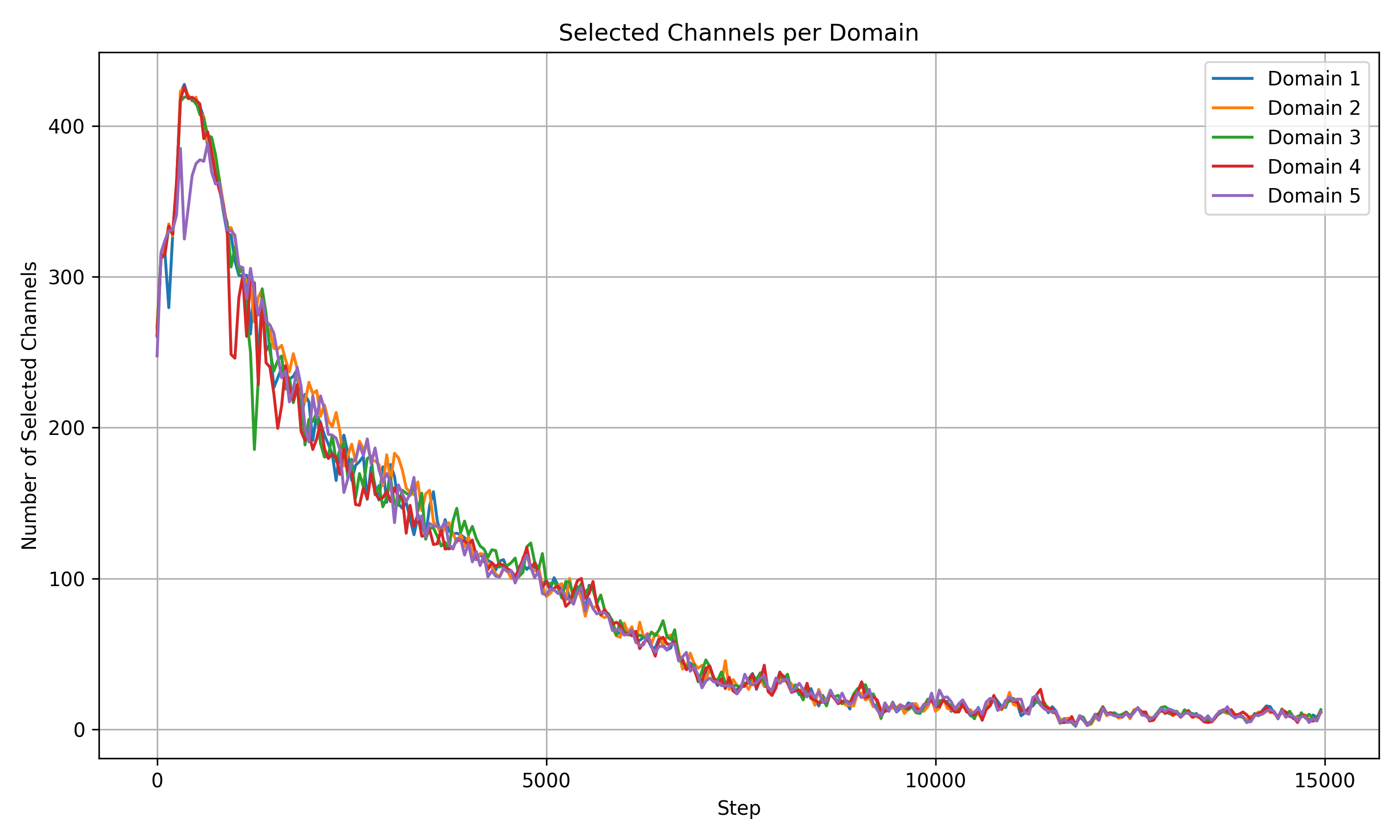}
        \caption{With consistency loss}
    \end{subfigure}
    \caption{Segmentation results with and without consistency loss.}
    \label{fig:consistency}
\end{figure}

\begin{figure*}[ht]
    \centerline{\includegraphics[width=0.75\textwidth]{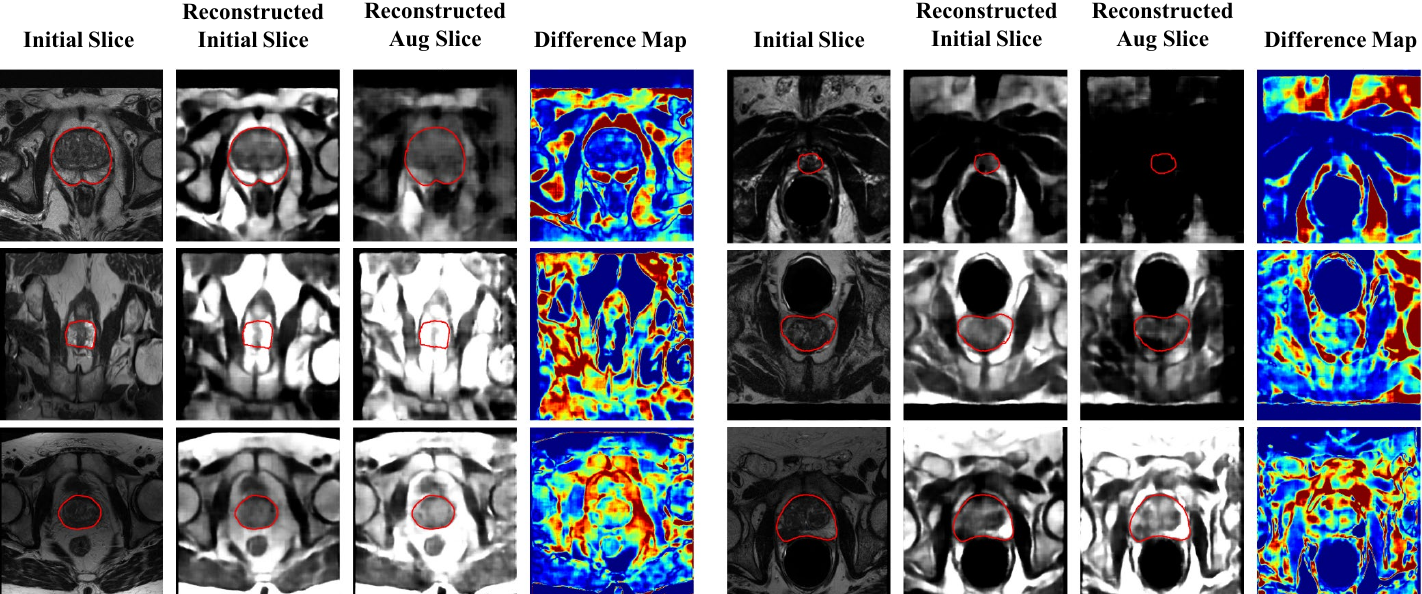}}
    \caption{Visualization results of semantically augmented features.}
    \label{diff map}
    \end{figure*}

\subsubsection{Impact of SCL}
To effectively validate the role of the SCL in stabilizing the training process and providing fine-grained regularization for the SDS, we conducted a comparative experiment with and without the SCL. By visualizing the number of channels selected throughout the training steps, we can intuitively observe the impact of the consistency constraint.

As shown in Figure~\ref{fig:consistency}, without the SCL, the number of selected channels fluctuates significantly during the early stages of training and remains unstable in the later stages. In contrast, when the SCL is applied, the SDS exhibits more stable behavior, selecting appropriate channels more smoothly in the early stages and maintaining consistency throughout the later training phase. This demonstrates that the SCL effectively suppresses unstable selection patterns and enhances the reliability of the feature selection process.

\subsubsection{Visualization of Semantic Direction}
To verify that SDS can achieve effective and controllable feature-level augmentation without compromising the anatomical structures of medical images, we design a lightweight reconstruction network for visualization. Specifically, we reconstruct the last-layer feature representations \(z\) output by the encoder of the segmentation network.  

As shown in Figure~\ref{diff map}, the first column shows the original slices, the second column shows the reconstructions from the original features, the third column shows the reconstructions from the augmented features, and the fourth column presents normalized pixel-wise difference maps. The reconstructed images are obtained by decoding the features back to the image space, followed by the inverse transformation of the image-level augmentations used during training to align with the original images and enable intuitive visualization.  

The difference maps indicate that the overall anatomical structures of the original images are well preserved after augmentation. Most changes occur in contextual regions surrounding the target areas, such as muscles, bones, and neighboring organs, while the target regions remain largely unchanged. The low intensity within the target regions suggests that the augmentation process respects their integrity, thus supporting reliable and robust segmentation. Overall, SDS introduces controlled perturbations mainly in domain-sensitive contextual regions while maintaining structural consistency.

\subsection{ablation study}
\subsubsection{Efficay of Each Module}

\begin{table}[htbp]
    \centering
    \caption{SDS = Semantic Direction Selector:, SIS = Semantic Intensity Sampler. The table shows DSC for each combination of modules in the A-F domain.}
    \begin{tabular}{llcccccccc}
    \toprule
    SDS & SIS &  A & B & C & D & E & F & Avg \\
    \midrule
     &  &  87.31 & 85.09 & 87.59 & 83.85 & 81.50 & 86.38 & 85.29 \\
     & \checkmark & 86.35 & \textbf{86.07} & 87.32 & 85.91 & 82.05 & 85.38 & 85.51 \\
     \checkmark &  & 87.41 & 85.07 & 88.07 & 85.47 & 80.43 & 87.34 & 85.63 \\
     \checkmark & \checkmark & \textbf{87.86} & 85.25 & \textbf{88.92} & \textbf{85.96} & \textbf{82.24} & \textbf{87.82} & \textbf{86.34} \\
    
    \bottomrule
    \end{tabular}
    \label{tab:ablation}
    \end{table}

We conducted systematic ablation experiments on the two key components of the proposed method, SDS and SIS, across six different domains (A--F). The experimental results are summarized in Table~\ref{tab:ablation}.  

In the baseline setting without any modules, channel selection was performed randomly following a Bernoulli distribution, and the enhancement intensity was sampled from a standard Gaussian distribution, yielding an average DSC of 85.29. After incorporating the SIS module, the average DSC increased to 85.51, indicating that more reasonable sampling of enhancement intensities can partially improve model performance. Introducing the SDS module alone further improved the average DSC to 85.63, demonstrating that the SDS can effectively identify feature channels suitable for augmentation compared with random selection.  

When both SDS and SIS modules were integrated, the model achieved the best performance across all domains, with an average DSC of 86.34, significantly outperforming all other combinations. This indicates strong complementarity between the two modules: SDS focuses on key feature dimensions, while SIS further enriches feature representations through controllable enhancement intensities, and their synergy drives consistent performance improvement.

\subsubsection{Efficay of SCL}
\begin{table}[htbp]
    \centering
    \caption{Performance Comparison With and Without Selective Consistency Loss (SCL)}
    \resizebox{\columnwidth}{!}{
    \begin{tabular}{llcccccccc}
    \toprule
    Method & A & B & C & D & E & F & Avg \\
    \midrule
    
    w/o SCL & 87.50 & 85.28 & 87.37 & 85.78 & \textbf{82.96} & 86.78 & 85.94 \\
    with SCL & \textbf{87.86} & 85.25 & \textbf{88.92} & \textbf{85.96} & 82.24 & \textbf{87.82} & \textbf{86.34} \\
    
    \bottomrule
    \end{tabular}
    }
    \label{tab:consistency_ablation}
    \end{table}

    To further validate the role of the SCL in enhancing the stability and fine-grained constraint of the SDS during training, we conducted an ablation study by comparing models trained with and without the SCL. As shown in Table~\ref{tab:consistency_ablation}, models incorporating the SCL achieved higher mDSC across the six domains, indicating its overall effectiveness.

\section{Conclusion}
In this work, we tackle the critical challenge of DG in medical image segmentation, where models must maintain high performance across unseen domains with diverse imaging conditions. We propose a domain-statistics-guided feature perturbation framework that enhances robustness while preserving essential anatomical semantics. Specifically, our approach integrates a Semantic Direction Selector (SDS) and a covariance-based Semantic Intensity Sampler (SIS) to enable adaptive and discriminative perturbations, and we introduce a Selective Consistency Loss (SCL) to stabilize training by selectively regularizing degraded augmented features. Extensive experiments on multi-center benchmarks demonstrate that our method consistently improves generalization, representing a promising step toward reliable and deployable segmentation models for real-world clinical scenarios.

\bibliographystyle{IEEEtran}  
\bibliography{bibm}     

\begin{thebibliography}{10}
\providecommand{\url}[1]{#1}
\csname url@samestyle\endcsname
\providecommand{\newblock}{\relax}
\providecommand{\bibinfo}[2]{#2}
\providecommand{\BIBentrySTDinterwordspacing}{\spaceskip=0pt\relax}
\providecommand{\BIBentryALTinterwordstretchfactor}{4}
\providecommand{\BIBentryALTinterwordspacing}{\spaceskip=\fontdimen2\font plus
\BIBentryALTinterwordstretchfactor\fontdimen3\font minus \fontdimen4\font\relax}
\providecommand{\BIBforeignlanguage}[2]{{%
\expandafter\ifx\csname l@#1\endcsname\relax
\typeout{** WARNING: IEEEtran.bst: No hyphenation pattern has been}%
\typeout{** loaded for the language `#1'. Using the pattern for}%
\typeout{** the default language instead.}%
\else
\language=\csname l@#1\endcsname
\fi
#2}}
\providecommand{\BIBdecl}{\relax}
\BIBdecl

\bibitem{litjens2017survey}
G.~Litjens, T.~Kooi, B.~E. Bejnordi, A.~A.~A. Setio, F.~Ciompi, M.~Ghafoorian, J.~A. Van Der~Laak, B.~Van~Ginneken, and C.~I. S{\'a}nchez, ``A survey on deep learning in medical image analysis,'' \emph{Medical image analysis}, vol.~42, pp. 60--88, 2017.

\bibitem{ronneberger2015u}
O.~Ronneberger, P.~Fischer, and T.~Brox, ``U-net: Convolutional networks for biomedical image segmentation,'' in \emph{Medical image computing and computer-assisted intervention--MICCAI 2015: 18th international conference, Munich, Germany, October 5-9, 2015, proceedings, part III 18}.\hskip 1em plus 0.5em minus 0.4em\relax Springer, 2015, pp. 234--241.

\bibitem{isensee2021nnu}
F.~Isensee, P.~F. Jaeger, S.~A. Kohl, J.~Petersen, and K.~H. Maier-Hein, ``nnu-net: a self-configuring method for deep learning-based biomedical image segmentation,'' \emph{Nature methods}, vol.~18, no.~2, pp. 203--211, 2021.

\bibitem{hatamizadeh2022unetr}
A.~Hatamizadeh, Y.~Tang, V.~Nath, D.~Yang, A.~Myronenko, B.~Landman, H.~R. Roth, and D.~Xu, ``Unetr: Transformers for 3d medical image segmentation,'' in \emph{Proceedings of the IEEE/CVF winter conference on applications of computer vision}, 2022, pp. 574--584.

\bibitem{hatamizadeh2021swin}
A.~Hatamizadeh, V.~Nath, Y.~Tang, D.~Yang, H.~R. Roth, and D.~Xu, ``Swin unetr: Swin transformers for semantic segmentation of brain tumors in mri images,'' in \emph{International MICCAI brainlesion workshop}.\hskip 1em plus 0.5em minus 0.4em\relax Springer, 2021, pp. 272--284.

\bibitem{glocker2019machine}
B.~Glocker, R.~Robinson, D.~C. Castro, Q.~Dou, and E.~Konukoglu, ``Machine learning with multi-site imaging data: An empirical study on the impact of scanner effects,'' \emph{arXiv preprint arXiv:1910.04597}, 2019.

\bibitem{yoon2024domain}
J.~S. Yoon, K.~Oh, Y.~Shin, M.~A. Mazurowski, and H.-I. Suk, ``Domain generalization for medical image analysis: A review,'' \emph{Proceedings of the IEEE}, 2024.

\bibitem{zhang2021prototypical}
P.~Zhang, B.~Zhang, T.~Zhang, D.~Chen, Y.~Wang, and F.~Wen, ``Prototypical pseudo label denoising and target structure learning for domain adaptive semantic segmentation,'' in \emph{Proceedings of the IEEE/CVF conference on computer vision and pattern recognition}, 2021, pp. 12\,414--12\,424.

\bibitem{tomar2021self}
D.~Tomar, M.~Lortkipanidze, G.~Vray, B.~Bozorgtabar, and J.-P. Thiran, ``Self-attentive spatial adaptive normalization for cross-modality domain adaptation,'' \emph{IEEE transactions on medical imaging}, vol.~40, no.~10, pp. 2926--2938, 2021.

\bibitem{dong2021and}
J.~Dong, Y.~Cong, G.~Sun, Z.~Fang, and Z.~Ding, ``Where and how to transfer: Knowledge aggregation-induced transferability perception for unsupervised domain adaptation,'' \emph{IEEE Transactions on Pattern Analysis and Machine Intelligence}, vol.~46, no.~3, pp. 1664--1681, 2021.

\bibitem{liu2020shape}
Q.~Liu, Q.~Dou, and P.-A. Heng, ``Shape-aware meta-learning for generalizing prostate mri segmentation to unseen domains,'' in \emph{International conference on medical image computing and computer-assisted intervention}.\hskip 1em plus 0.5em minus 0.4em\relax Springer, 2020, pp. 475--485.

\bibitem{ouyang2022causality}
C.~Ouyang, C.~Chen, S.~Li, Z.~Li, C.~Qin, W.~Bai, and D.~Rueckert, ``Causality-inspired single-source domain generalization for medical image segmentation,'' \emph{IEEE Transactions on Medical Imaging}, vol.~42, no.~4, pp. 1095--1106, 2022.

\bibitem{gao2023bayeseg}
S.~Gao, H.~Zhou, Y.~Gao, and X.~Zhuang, ``Bayeseg: Bayesian modeling for medical image segmentation with interpretable generalizability,'' \emph{Medical Image Analysis}, vol.~89, p. 102889, 2023.

\bibitem{zhu2025improving}
J.~Zhu, B.~Bolsterlee, Y.~Song, and E.~Meijering, ``Improving cross-domain generalizability of medical image segmentation using uncertainty and shape-aware continual test-time domain adaptation,'' \emph{Medical Image Analysis}, vol. 101, p. 103422, 2025.

\bibitem{billot2023synthseg}
B.~Billot, D.~N. Greve, O.~Puonti, A.~Thielscher, K.~Van~Leemput, B.~Fischl, A.~V. Dalca, J.~E. Iglesias \emph{et~al.}, ``Synthseg: Segmentation of brain mri scans of any contrast and resolution without retraining,'' \emph{Medical image analysis}, vol.~86, p. 102789, 2023.

\bibitem{shen2022randstainna}
Y.~Shen, Y.~Luo, D.~Shen, and J.~Ke, ``Randstainna: Learning stain-agnostic features from histology slides by bridging stain augmentation and normalization,'' in \emph{International Conference on Medical Image Computing and Computer-Assisted Intervention}.\hskip 1em plus 0.5em minus 0.4em\relax Springer, 2022, pp. 212--221.

\bibitem{scalbert2022test}
M.~Scalbert, M.~Vakalopoulou, and F.~Couzini{\'e}-Devy, ``Test-time image-to-image translation ensembling improves out-of-distribution generalization in histopathology,'' in \emph{International Conference on Medical Image Computing and Computer-Assisted Intervention}.\hskip 1em plus 0.5em minus 0.4em\relax Springer, 2022, pp. 120--129.

\bibitem{meng2020mutual}
Q.~Meng, J.~Matthew, V.~A. Zimmer, A.~Gomez, D.~F. Lloyd, D.~Rueckert, and B.~Kainz, ``Mutual information-based disentangled neural networks for classifying unseen categories in different domains: Application to fetal ultrasound imaging,'' \emph{IEEE transactions on medical imaging}, vol.~40, no.~2, pp. 722--734, 2020.

\bibitem{bi2023mi}
Y.~Bi, Z.~Jiang, R.~Clarenbach, R.~Ghotbi, A.~Karlas, and N.~Navab, ``Mi-segnet: Mutual information-based us segmentation for unseen domain generalization,'' in \emph{International Conference on Medical Image Computing and Computer-Assisted Intervention}.\hskip 1em plus 0.5em minus 0.4em\relax Springer, 2023, pp. 130--140.

\bibitem{zhou2022generalizable}
Z.~Zhou, L.~Qi, X.~Yang, D.~Ni, and Y.~Shi, ``Generalizable cross-modality medical image segmentation via style augmentation and dual normalization,'' in \emph{Proceedings of the IEEE/CVF conference on computer vision and pattern recognition}, 2022, pp. 20\,856--20\,865.

\bibitem{chen2023treasure}
Z.~Chen, Y.~Pan, Y.~Ye, H.~Cui, and Y.~Xia, ``Treasure in distribution: A domain randomization based multi-source domain generalization for 2d medical image segmentation,'' in \emph{International Conference on Medical Image Computing and Computer-Assisted Intervention}.\hskip 1em plus 0.5em minus 0.4em\relax Springer, 2023, pp. 89--99.

\bibitem{zhang2024domain}
Z.~Zhang, B.~Wang, D.~Jha, U.~Demir, and U.~Bagci, ``Domain generalization with correlated style uncertainty,'' in \emph{Proceedings of the IEEE/CVF Winter Conference on Applications of Computer Vision}, 2024, pp. 2000--2009.

\bibitem{upchurch2017deep}
P.~Upchurch, J.~Gardner, G.~Pleiss, R.~Pless, N.~Snavely, K.~Bala, and K.~Weinberger, ``Deep feature interpolation for image content changes,'' in \emph{Proceedings of the IEEE conference on computer vision and pattern recognition}, 2017, pp. 7064--7073.

\bibitem{wang2021regularizing}
Y.~Wang, G.~Huang, S.~Song, X.~Pan, Y.~Xia, and C.~Wu, ``Regularizing deep networks with semantic data augmentation,'' \emph{IEEE Transactions on Pattern Analysis and Machine Intelligence}, vol.~44, no.~7, pp. 3733--3748, 2021.

\bibitem{zhu2024bsda}
Y.~Zhu, X.~Cai, X.~Wang, X.~Chen, Z.~Fu, and Y.~Yao, ``Bsda: Bayesian random semantic data augmentation for medical image classification,'' \emph{Sensors}, vol.~24, no.~23, p. 7511, 2024.

\bibitem{wang2020dofe}
S.~Wang, L.~Yu, K.~Li, X.~Yang, C.-W. Fu, and P.-A. Heng, ``Dofe: Domain-oriented feature embedding for generalizable fundus image segmentation on unseen datasets,'' \emph{IEEE Transactions on Medical Imaging}, vol.~39, no.~12, pp. 4237--4248, 2020.

\bibitem{bi2024learning}
Q.~Bi, J.~Yi, H.~Zheng, W.~Ji, Y.~Huang, Y.~Li, and Y.~Zheng, ``Learning generalized medical image segmentation from decoupled feature queries,'' in \emph{Proceedings of the AAAI conference on artificial intelligence}, vol.~38, no.~2, 2024, pp. 810--818.

\bibitem{sivaswamy2015comprehensive}
J.~Sivaswamy, S.~Krishnadas, A.~Chakravarty, G.~Joshi, A.~S. Tabish \emph{et~al.}, ``A comprehensive retinal image dataset for the assessment of glaucoma from the optic nerve head analysis,'' \emph{JSM Biomedical Imaging Data Papers}, vol.~2, no.~1, p. 1004, 2015.

\bibitem{fumero2011rim}
F.~Fumero, S.~Alay{\'o}n, J.~L. Sanchez, J.~Sigut, and M.~Gonzalez-Hernandez, ``Rim-one: An open retinal image database for optic nerve evaluation,'' in \emph{2011 24th international symposium on computer-based medical systems (CBMS)}.\hskip 1em plus 0.5em minus 0.4em\relax IEEE, 2011, pp. 1--6.

\bibitem{orlando2020refuge}
J.~I. Orlando, H.~Fu, J.~B. Breda, K.~Van~Keer, D.~R. Bathula, A.~Diaz-Pinto, R.~Fang, P.-A. Heng, J.~Kim, J.~Lee \emph{et~al.}, ``Refuge challenge: A unified framework for evaluating automated methods for glaucoma assessment from fundus photographs,'' \emph{Medical image analysis}, vol.~59, p. 101570, 2020.

\bibitem{zhou2021domain}
K.~Zhou, Y.~Yang, Y.~Qiao, and T.~Xiang, ``Domain generalization with mixstyle,'' \emph{arXiv preprint arXiv:2104.02008}, 2021.

\bibitem{zhang2022exact}
Y.~Zhang, M.~Li, R.~Li, K.~Jia, and L.~Zhang, ``Exact feature distribution matching for arbitrary style transfer and domain generalization,'' in \emph{Proceedings of the IEEE/CVF conference on computer vision and pattern recognition}, 2022, pp. 8035--8045.

\end{thebibliography}

\end{document}